\documentclass[10pt,twocolumn,letterpaper]{article}

\usepackage{3dv}
\usepackage{times}
\usepackage{epsfig}
\usepackage{graphicx}
\usepackage{amsmath}
\usepackage{amssymb}

\usepackage{ctable}
\usepackage[normalem]{ulem}
\useunder{\uline}{\ul}{}
\usepackage{wrapfig}

\usepackage[pagebackref=true,breaklinks=true,colorlinks,bookmarks=false]{hyperref}

\newcommand{\todo}[1]{\textcolor{red}{TODO:#1}}

\newcommand{\old}[1]{\textcolor{red}{}}

\newcommand{\AgHSRep}[1]{16}
\newcommand{\AgSSRep}[1]{17}

\threedvfinalcopy % *** Uncomment this line for the final submission

\def\threedvPaperID{302} % *** Enter the 3DV Paper ID here

\ifthreedvfinal\fi
\begin{document}

%%%%%%%%% TITLE
\title{\vspace{-10pt}Fine-Grained VR Sketching: Dataset and Insights.  }

\author{Ling Luo$^{1,2}$
\and 
Yulia Gryaditskaya$^{1,2}$
\and 
Yongxin Yang$^{1,2}$
\and
Tao Xiang$^{1,2}$
\and
Yi-Zhe Song$^{1,2}$
\and\\
$^{1}$SketchX, CVSSP, University of Surrey $^{2}$iFlyTek-Surrey Joint Research Centre on Artificial Intelligence\\
}

% \and

\maketitle

\thispagestyle{empty}

%%%%%%%%% ABSTRACT
\begin{abstract}
We present the first fine-grained dataset of 1,497 3D VR sketch and 3D shape pairs of a chair category with large shapes diversity. Our dataset supports the recent trend in the sketch community on fine-grained data analysis, and extends it to an actively developing 3D domain. We argue for the most convenient sketching scenario where the sketch consists of sparse lines and does not require any sketching skills, prior training or time-consuming accurate drawing.
We then, for the first time, study the scenario of fine-grained 3D VR sketch to 3D shape retrieval, as a novel VR sketching application and a proving ground to drive out generic insights to inform future research.
By experimenting with carefully selected combinations of design factors on this new problem, we draw important conclusions to help follow-on work. We hope our dataset will enable other novel applications, especially those that require a fine-grained angle such as fine-grained 3D shape reconstruction. 
The dataset is available at \url{tinyurl.com/VRSketch3DV21}.
\end{abstract}

\vspace{-10pt}
\section{Introduction}
Virtual Reality (VR) headsets and 3D printers rapidly make their way to consumer markets.
With recent interest in virtual reality, VR sketching is gaining increasingly more popularity in industry\footnote{https://www.gravitysketch.com/, https://www.tiltbrush.com/ and https://coolpaintrvr.com/en/} and academia \cite{rosales2019surfacebrush,oti2020immersive,joundi2020explorative, yang2020cognitive,yu2021cassie,arora2021mid}. 
In this work, we investigate the potential of \emph{low-effort} VR sketching to become a bridge to the practical adoption of 3D and VR-related technologies by average consumers and professional designers.

In sketching research, the recent focus is on fine-grained tasks revolving around subtle intra-class differences \cite{yu2020fine, bhunia2020sketch, pang2020solving, zhong2020deep, zhong2020towards}. 
In particular, 2D sketches were proved to be efficient queries for 2D images fine-grained retrieval \cite{shao2011discriminative, yu2016sketch, yu2020fine, dutta2020semantically}.
Yet, in the context of the 3D shape retrieval from single or multiple 2D sketches the fine-grained performance was not demonstrated so far \cite{eitz2012sketch, wang2015sketch, zhu2016learning, xu2018sketch, jiao2020cross}.

\begin{figure}[t]
\vspace{-8pt}
\begin{center}
%\fbox{\rule{0pt}{3in} \rule{1.0\linewidth}{0pt}}
 \includegraphics[width=\columnwidth]{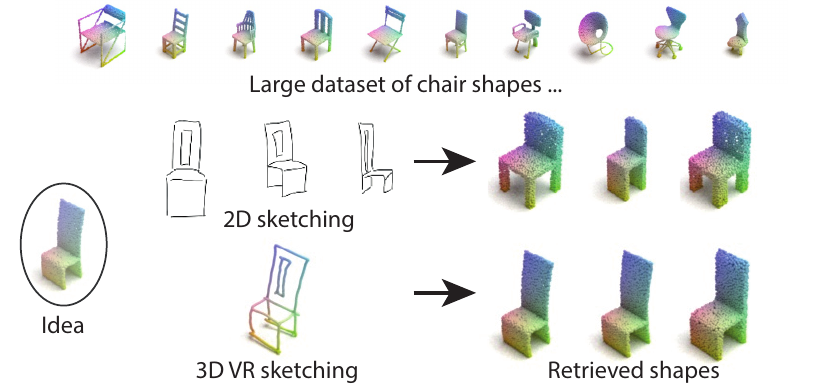}
\end{center}
\vspace{-8pt}
   \caption{\old{We collect a dataset of 1,497 human sketches of diverse chairs, and study its properties on one proving ground application: Searching for an envisioned shape can be frustrating and time-consuming, we argue for a new modality for a fine-grained retrieval and generation -- fast 3D VR sketch. 
\todo{Change to motivation of 3D VR sketching compared to 3D sketching, change figure and text.}}
%   We argue for a new modality for a fine-grained retrieval and generation -- fast 3D VR sketch. \todo{Change teaser}
    We investigate the potential of low-effort VR sketching as the enabler of the practical adoption of 3D and VR-related technologies and study \emph{fine-grained} 3D sketch to 3D shape retrieval.}
\label{fig:teaser}
\vspace{-14pt}
\end{figure}

2D sketches contain a 2D projection ambiguity, where multiple 3D shapes can project to the same 2D sketch \cite{xu2014true2form}, while sketching the same shape from multiple 2D viewpoints is a non-trivial task. 
Furthermore, even professionals find it challenging to accurately depict shape proportions and dimensions in a 2D sketch \cite{schmidt2009expert}.
\cite{oti2021immersive} conclude that sketching in 3D gives better shape understanding.
3D sketching allows to (1) alleviate the problem of a 2D projection ambiguity and (2) naturally evaluate and depict shapes' proportions (Fig.~\ref{fig:teaser}).
Driven by these observations, we study, for the first time, the problem of \emph{fine-grained} intra-category 3D sketch to 3D shape retrieval.
%with low-effort VR sketches as queries.  

Little work thus far was done on 3D sketch to 3D shape retrieval, exploiting either a small set of inaccurate sketches collected with Microsoft Kinect \cite{li2016shrec, ye20163d}, or dense colorful sketches drawn on top of the initial 3D model\cite{giunchi20183d}. Research on free-hand VR sketches is hindered by a lack of training data.
As a first step, Luo et al.~\cite{Luo2020VRSketch} collected a small dataset of 167 sketches for testing, and proposed a heuristic method to generate 3D sketches with different abstractness levels. They were the first to study the retrieval problem from sparse human VR sketches, and showed that a reasonable \emph{inter-class} retrieval accuracy on human sketches can be achieved by training on synthetic sketches, even though inferior to the accuracy on synthetic data. Nevertheless, the \emph{intra-class} Top-1 accuracy of their method is low. We aim at increasing intra-class accuracy, enabling the retrieval of a particular instance of a shape within a single category (fine-grained). We meticulously evaluate alternative network designs, comparing several state-of-the-art encoders \cite{qi2017pointnet++,wang2019dynamic} and losses \cite{wang2014learning,khosla2020supervised}.

While in the absence of human sketches, synthetic sketches are the only option, the method for synthetic sketch generation proposed in \cite{Luo2020VRSketch} requires an input mesh to be a clean manifold curvature-aligned quad-dominant mesh. For new categories, collecting human sketches might be more feasible than obtaining 3D shapes with required properties. However, collecting sketch datasets is a labor-intensive task \cite{eitz2012humans, ha2017neural, zou2018sketchyscene, gryaditskaya2019opensketch}.
We therefore focus on a single chair category and collect 1,497 fine-grained 3D VR sketch and shape pairs, which we use to drive out insights on the desirable properties of 3D VR sketch datasets for fine-grained tasks. We select the chair category due to the large variability of shapes, dimensions, and details, including shapes with variable genus values, shapes dominated by planar or non-planar surfaces (Fig.~\ref{fig:sketches_learning}). Such large intra-class variability makes this category ideal for our fine-grained goal.
We perform exhaustive evaluations on (i) how the dataset size and its composition affect retrieval accuracy, (ii) what the expected performance gain from training on human sketches compared to synthetic data is, (iii) how human and synthetic sketches can complement each other, (iv) the role of data augmentation, (v) how human ability to sketch in 3D increases over time, and (vi) how sketching style differences affect the performance. 

Targeting practical novel applications, we assert a number of guiding principles to the overall design of our dataset: (P1) \emph{convenience}: we argue for the most convenient sketching scenario where the sketch consists of sparse lines and does not require any sketching skills, prior training or time-consuming accurate drawing; 
(P2) \emph{fine-grained sketching}: we require sketches to be fine-grained, capturing salient visual details of a \textit{paired} 3D shape, that sufficiently differentiates it from other 3D sketches within the same category; 
(P3) \emph{free-space sketching}: we require sketches to faithfully reflect humans' ability to sketch in free space;
and (P4) \emph{diversity}: we require sketches to contain a diversity of styles and levels of details. 
We capture the sketching process, recording the time-space coordinates of each stroke. 

In summary, our  contributions  include: (i) the first large-scale dataset of paired human 3D VR sketches and 3D shapes, 
(ii) key insights associated with the dataset to inform future work on 3D VR sketches, 
(iii) the first in-depth study of a proving ground application: fine-grained 3D sketch based 3D shape retrieval, and conclude with (iv) a series of key technical insights to guide future research.

\section{Related work}
\subsection{Sketching datasets}
When collecting 2D sketches by novices it is possible to rely on crowd-sourcing and readily available hardware \cite{eitz2012humans, ha2017neural, zou2018sketchyscene,sangkloy2016sketchy}. 
Our task is more challenging since VR headsets are not yet commonly available, and that VR sketching is still a relatively novel concept with limited off-the-shelf tools available.
%Taking this into account, we take careful consideration in designing our dataset. 
The earlier works on sketching were targeting category-level sketch understanding, such as the TUBerlin dataset \cite{eitz2012humans} which consists of 20,000 sketches in total but only 80 sketches per category. 
Recent work focuses on fine-grained tasks, offering datasets consisting of 1-2 categories with a larger number of sketches. \Eg, \cite{yu2016sketch} proposed a dataset of two categories, shoes and \emph{chairs}, with 419 and 297 sketch-photo pairs, accordingly.  \cite{zhong2020towards} proposed a dataset of 1,500 professional 2D sketches collected across 500 3D chair shapes. 
%Similarly, recent dataset AmateurSketch-3DChair\footnote{http://sketchx.ai/downloads/: AmateurSketch-3DChair} consists of 3 2D views by novices of 1,005 3D shapes. 
Similarly, recent dataset AmateurSketch-3DChair\footnote{http://sketchx.ai/downloads/: AmateurSketch-3DChair} consists of 3 2D views by novices of 1,005 3D shapes. 
% Following this trend, and to enable the comparison with 2D sketch-based retrieval, we collect the 1,497 3D VR sketches for the same 1,005 shapes, with 492 shapes sketched by two participants. 
Following this trend, and to enable the comparison with 2D sketch-based retrieval, we collect the 1,497 3D VR sketches for the same 1,005 shapes. 492 shapes are sketched by two participants each.

% Recently, Gryaditskaya et al.~\cite{gryaditskaya2019opensketch} collected a dataset of professional sketches. While such task of collecting professional sketches is more challenging than the ones by novices, they were able to deploy a custom-build interface, which runs on a tablet -- a tool commonly used by professionals. 
% Our task is much more challenging, since VR headsets are not yet commonly available, and that VR sketching is still a relatively novel concept with limited off-the-shelf tools available.
%\todo{needs expanding a bit to spell out the difficulties/differences building our dataset}

\subsection{Retrieval}
\paragraph{2D sketch/image to 3D shape/ 2D image}
%The cross-entropy loss is common for classification task and can be used for category-level retrieval. \todo{UNCLEAR, what is the first part of this sentence trying to say? More accurate retrieval results can be obtained with contrastive losses, yet triplet loss and its variants \cite{ wang2014learning, schroff2015facenet, wang2015sketch, yu2016sketch, wen2016discriminative, collomosse2017sketching, grabner20183d, yu2020fine,song2017deep} are a common choice. }

Triplet loss and its variants \cite{ wang2014learning, schroff2015facenet, wang2015sketch, yu2016sketch, wen2016discriminative, collomosse2017sketching, grabner20183d, yu2020fine,song2017deep} are by far the most commonly used losses for the retrieval task.
% Triplet loss combined with attention modules were shown to result in higher accuracy of the sketch to image retrieval \cite{song2017deep, yu2020fine}.
A number of recent losses \cite{liu2017rethinking, li2019angular,li2019rethinking} were shown to improve category-level retrieval performance, but are less suitable for fine-grained retrieval.
We explore here a very recently proposed contrastive loss by Khosla et al~\cite{khosla2020supervised}.
%that operates on cosine based distances on the normalized feature space.
Compared to the triplet loss, their formulation allows to have multiple positive and negative samples and was shown to better learn inter- and intra-class variability. 
%It is also shown to be not so sensitive to the hyper-parameters.

\vspace{-8pt}
\paragraph{3D sketch to 3D shape}
3D sketch to 3D shape retrieval is a relatively novel field \cite{li20153d, li2016shrec, ye20163d, giunchi20183d, Luo2020VRSketch}. The progress is hampered by a lack of datasets of human 3D sketches. 
Most of the existing works \cite{li20153d, li2016shrec, ye20163d} relied on the dataset collected using Microsoft Kinect, which has limited tracking accuracy. As a result, these sketches have low fidelity and exhibit less details, making it impractical for fine-grained methods. Moreover, this dataset is no longer accessible online. 
Luo et al.~\cite{Luo2020VRSketch} addressed the category-level retrieval on sparse 3D VR sketches created by novices. They proposed a synthetic data generation method from 3D shapes and collected a small dataset of human sketches that they use as test data. 
%Despite being able to generate the data with different levels of abstractness, they showed that the performance on human sketches drops compared to the performance on synthetic data. 
They showed that the performance on human sketches drops compared to the performance on synthetic data. Given this drop, here we study how training on human sketches affects the retrieval performance, and how it compares with training on synthetic sketches.
Unlike \cite{Luo2020VRSketch} we focus on the fine-grained retrieval and drive out insights on the desirable structure of 3D VR  sketch datasets.

%\todo{ In this work we aim at studying the potential of deep 3D sketch to 3D shape retrieval methods, but using a fine-grained setting where retrieval is conducted intra-category other than inter-category as in their work. } 

%when the domain gap between the training and test data is minimized.
% To this end we collect a set of 1005 3D VR sketches of chairs. While we are the first to collect the dataset of 3D VR sketches of such scale, due to be extremely labor-intensive task it covers only 1/6th of the full set of shapes in ShapeNetCore \cite{shapenet}. We thus propose and evaluate deep data augmentation strategies, that aim at generating synthetic sketches matching the human sketches, without a need for heavy mesh processing algorithms.

\vspace{-8pt}
\paragraph{3D shape to 3D shape} Closely related to our problem is 3D shape-based 3D shape retrieval.
In 3D shape to 3D shape retrieval it is common to represent shapes via multi-view projections \cite{liu2018cross, he2018triplet,he2019view, li2019angular}. The 3D VR sketches we are dealing with are sparse, and, as it was shown in \cite{Luo2020VRSketch},  multi-view representations lose to point cloud based ones on a problem similar to ours. Deng et al.~\cite{deng2018ppfnet} exploit point pair features, based on the normal vector values, to align the point clouds. Such an approach can not be efficiently applied to 3D sketches, since the normal to a line is not uniquely defined in 3D, and the notion of patch is not well defined on a sparse sketch. 
Uy et al.~\cite{uy2020deformation} proposed a novel framework for deformation-aware 3D retrieval based on point cloud shape representation. It is not straightforward to extend this work to 3D sketches, since it requires ensuring the deformation of a 3D shape is consistent with a \emph{sparse} sketch.
Dahnert et al.~\cite{dahnert2019joint} studied fine-grained retrieval from 3D scans to CAD models. Similarly to previous works, they train with the triplet loss, which we also leverage in our work. 
%The novelty comes in two additional networks, where the first is trained to segment the 3D scan to background and foreground and the second encoder-decoder architecture aims at foreground shape completion.

%% Point cloud networks
\subsection{Point cloud networks}
In our work we represent sparse VR sketches as point clouds. Point cloud shape representation received a lot of attention in recent years, being a natural data format for 3D scans. Many works were proposed for representation learning, and un/conditional point clouds generation. They can be divided into those based on encoder-decoder architectures \cite{fan2017point,achlioptas2018learning,yang2018foldingnet,sun2020pointgrow}, those with adversarial training \cite{gadelha2017shape,achlioptas2018learning,li2018point, gadelha2018multiresolution, Shu_2019_ICCV, jiang2018gal, valsesia2018learning, wang2020rethinking}, and those that exploit normalizing flows \cite{yang2019pointflow,pumarola2020c, kim2020softflow}. 
Some works address point clouds completion and denoising \cite{dai2017complete, dai2018scancomplete, yuan2018pcn, li2019pu, dahnert2019joint, avetisyan2019scan2cad, wang2020cascaded, rakotosaona2020pointcleannet}. Other works target learning shape properties \cite{guerrero2018pcpnet, ben2019nesti, pistilli2020point, lu2020deep}. 
The most relevant to us are the works on point clouds encoding \cite{ravanbakhsh2016deep, qi2017pointnet, qi2017pointnet++, atzmon2018point, wang2019dynamic, li2018pointcnn}.
% Ravanbakhsh et al.~\cite{ravanbakhsh2016deep} introduced a permutation-equivariant layer that allows to treat the input as a set rather than a vector.
We evaluate in our work the two most commonly used encoders: \emph{PointNet++} \cite{qi2017pointnet++} and \emph{DGCNN} \cite{wang2019dynamic}.

\section{Data collection}
% We are the first to collect a large-scale dataset of 3D VR sketches, some example sketches are shown in Figure \ref{fig:sketches_split}.

\paragraph{Interface \&  Task}

\begin{wrapfigure}[5]{r}{3.5cm}
\vspace{-0.2in}
\hspace{-0.2in}
\includegraphics[width = 3.8cm]{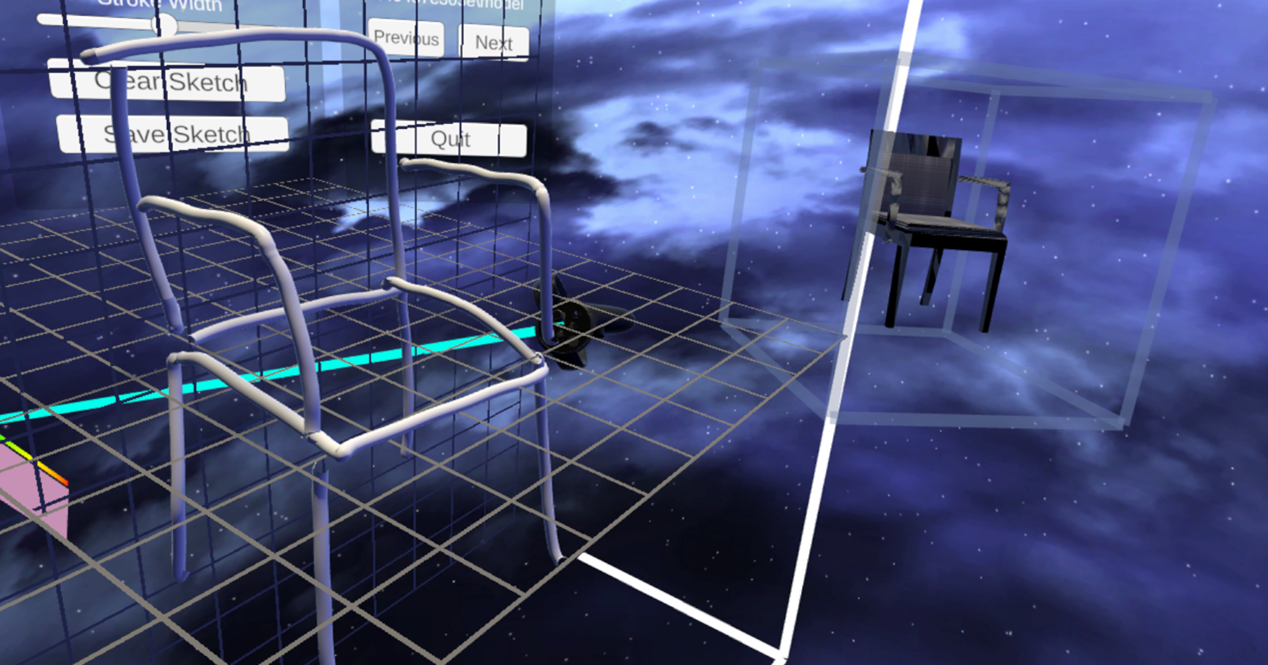}
\hspace{-0.1in}
\end{wrapfigure}
We base our data-collection interface on the one proposed in \cite{Luo2020VRSketch}\footnote{https://tinyurl.com/3DSketch3DV}, where participants were sketching over the reference shape. 
In our work, to better capture accuracy of sketching in free space, the reference is shown in 3D in an area separate from the sketching area, where 3D shape can be freely rotated, as shown in the inset. 
To improve space perception we displayed guiding grids \cite{arora2017experimental}.

Compared to \cite{Luo2020VRSketch}, we let participants choose line width, best matching the intention of the sketcher: thinner lines allow to create more detailed sketches, while thicker lines can be used to quickly get a general shape. This allows us to better capture the diversity of human styles.

To get familiar with sketching in 3D all the participants are asked to go through the following steps: (1) get familiar with hand controllers and the functions of each button and trigger: sketch, undo last stroke, delete all strokes, menu click button; (2) practice grabbing and rotating reference space and sketching space; (3) practice adjusting line width (4) practice drawing random lines. There was no training on how to sketch in 3D.

\vspace{-8pt}
\paragraph{Shapes}
We collected 1,497 sketches for a subset of 1,005 3D chairs in ShapeNetCore \cite{shapenet}. The chairs category is chosen due to a large intra-class shapes variability.
%, contributing to the fine-grained aspect of our dataset. 
The selected subset is representative of the full set of shapes. 

\begin{figure}[t]
\vspace{-4pt}
\centering
\begin{center}
%\fbox{\rule{0pt}{3in} \rule{1.0\linewidth}{0pt}}
 \includegraphics[width=1.02\linewidth]{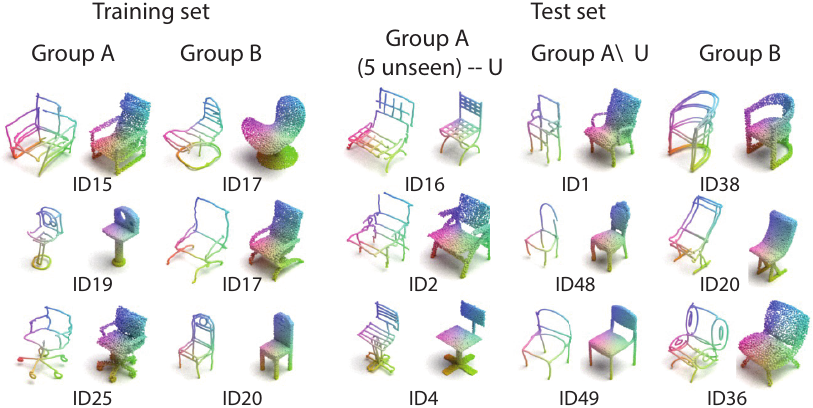}
\end{center}
\vspace{-10pt}
 \caption{Example sketch-shape pairs from the training and test sets. Sketches in our dataset cover large variability of shapes and vary in amount of details and accuracy. Since the participants from group $B$ (Sec.~\ref{sec:data_split}) draw more shapes than the participants from group $A$, their sketches tend to be more accurate and contain more details. It is interesting to observe how participants depict solid surfaces, some draw only ridge lines and some draw additional sets of parallel lines lying on the surface of a shape. Sometimes both types of depictions can be observed in the sketches of the same participant, e.g. the sketches of the participant 17 (ID17). }
\label{fig:sketches_split}
\vspace{-6pt}
\end{figure}

\begin{figure*}[t]
\vspace{-8pt}
\begin{center}
 \includegraphics[width=\linewidth]{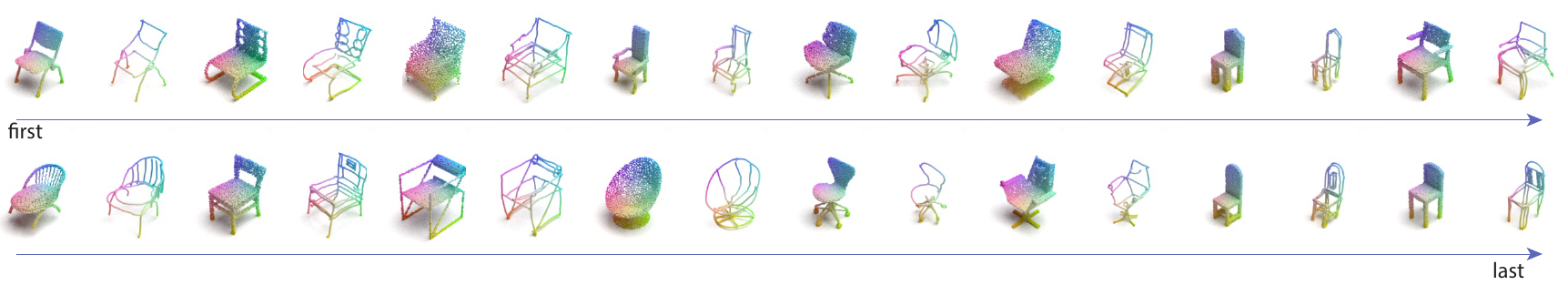}
\end{center}
\vspace{-14pt}
   \caption{We selected the sketches by the participant 20, evenly spaced in time of creation. This participant sketched the largest amount of shapes. The figure shows shape-sketch pairs, ordered by time of creation: from first to last. It can be seen that the later sketches in overall better capture the proportions of the shapes and contain more details than the first ones.}
\label{fig:sketches_learning}
\vspace{-10pt}
\end{figure*}

\vspace{-8pt}
\paragraph{Participants}
We split participant into two groups (Fig.~\ref{fig:sketches_split}).

\emph{Group A}: We hired $46$ participants who sketched 10 non-overlapping shapes each.
The number of shapes was limited for each participant to avoid fatigue and to limit the learning effect on sketching style. 

\emph{Group B}:
%We hired $50$ participants.
%, and assigned to them non-overlapping subsets of 1,005 shapes.  
This group consists of 4 participants, who sketched more. They sketched: 21, 72, 330 and 120 unique shapes. The last participant additionally sketched 492 shapes overlapping with the shapes from group A.
Sketches from these participants allow us to study how sketching style evolved over time, and how it affects retrieval results.
For instance, Fig.~\ref{fig:sketches_learning} shows, on an example of one participant, that the sketches are getting more accurate and detailed over time. 
%It can be seen in Figure \ref{fig:sketches_learning}, on example of one of the participants, that the sketches are getting more accurate and more detailed with time. 

At the end of the sketching task, the participants were asked to complete a questionnaire. The age of the participants varies between 18 and 40 years old. The average sketching experience on a Likert scale is 1.84, where 1 means no sketching experience and 5 means professional. 20 of the participant reported that they usually do not sketch, 23 do sketches several times a year, 5 several times a month and 2 several times per week. 33 participants had no prior experience with VR. The average score of how comfortable the participants felt sketching in our interface was 3.44, where 1 means not comfortable and 5 means comfortable. Being above average it indicates that even inexperienced participants found sketching in 3D satisfying. 
% The median sketching time for one shape is 2 min. 21 sec. \todo{Check other statistics on sketching time, and time statistics per group}
%Finally, we asked the participants how close their sketch is to what they had in mind. The participants average score is 2.54, where 1 means recognizable and 5 means matches perfectly. 

\vspace{-8pt}
\paragraph{Separation into train, test and validation.} \label{sec:data_split}
In our experiments, unless specified otherwise, we use only 1,005 sketches of unique shapes.
To split the collected data into three sets, we first held all the sketches of five randomly selected participants from group A as a test set to check that the compared methods generalize well to unseen human styles. For the rest of participants we split their sketches of unique shapes proportionally between the training, validation and test sets with the ratios $7:1:2$. 
In total, the training set consists of 702 sketches, the validation set of 101 sketches, and the test set of 202 sketches. We exploit the validation set to choose networks' hyper-parameters. See Fig.~\ref{fig:sketches_split} for visualization of the sketches from test and training sets. 
% \todo{maybe add how we use the rest of sketches}

\section{Retrieval}
Following the analysis from \cite{Luo2020VRSketch} we select the point cloud representation for 3D sketches and shapes. To train for \emph{fine-grained} retrieval, we compare a traditionally used triplet loss \cite{wang2014learning}, and a more recent contrastive loss \cite{henaff2019data,khosla2020supervised}.

\subsection{Data preparation \& augmentation}
\label{subsec:data}
All the target shapes in the ShapeNet dataset are oriented consistently. 
Since the sketches were drawn in the free-space, guided only by a sparse grid, the sketches' global orientations can be arbitrary, but are aligned with the grid axes.
In the collected dataset 68 sketches have vertical rotation (z-axis) inconsistent with the orientation of the 3D shapes, and 19 have x/y-axis rotations. 
We experiment with two versions of the dataset: (1) where we keep the original rotation of sketches and (2) where the sketches are manually rotated to match the orientation of the 3D shapes. All the shapes and sketches are centered and normalized to have a maximum dimension along x-,y-,z-axes to be equal to 1. 

Since collecting human sketches is a labor-intensive and time-consuming task, we experiment with several augmentation strategies. 
First, we apply randomly selected rotations in the range [0, 360] degrees around the vertical axis to the human sketches. 
Second, we apply global distortions by scaling sketches along 3 axes with 3 scale factors in a range [0.9, 1.1]. Both augmentations are applied to input on the fly when training.
Finally, we complement humans sketches with synthetic sketches, generated as was proposed in \cite{Luo2020VRSketch} with the abstractness level $1.0$. This setting was shown to be one of the optimal ones when tested on human sketches. 

\subsection{Evaluated losses}

\paragraph{Triplet loss}
Triplet loss ensures that the anchor-negative distances are larger than the anchor-positive distances by a given margin. Let $I$ be a set of indices of all objects in the given mini-batch. For each object we have the corresponding sketch and shape point clouds.
The triplet loss is a sum over all the triplets within a mini-batch:
\begin{equation}
    L_{TL} = \sum_{a \in I}\max\{0, ||s^a - z_{pos}^a||_2^2 - ||s^a - z_{neg}^{b}||_2^2 + m\},
    %L_{TL} = \sum_{i}\max\{0, d_{pos}^i - d_{neg}^i + m\},
    \label{math:triplet}
\end{equation}
where $||\cdot||_2$ is the Euclidean distance, and the feature space is normalized to a unit hypersphere.
We use an embedding $s^a$ of a 3D sketch as an anchor. We use an embedding $z_{pos}^a$ of the 3D shape, that sketch corresponds to, as a positive example.
A negative example can be an embedding $z_{neg}^{b}$ of any other shape in the batch, $b \in I \setminus a$.
We set the margin $m$ to $0.3$, a common choice for a fine-grained retrieval task. We also found this value to be optimal on our validation set. 

%When we use data augmentation, for each sketch we create one additional augmented sketch. We use the original and the augmented sketches as anchors, corresponding shape as a positive examples, and the rest of shapes as negative examples.

\vspace{-8pt}
\paragraph{Contrastive loss}
Contrastive losses proposed by \cite{oord2018representation} and \cite{khosla2020supervised} pursue the same goal as a triplet loss, but its formulation allows multiple negatives in the first case, and multiple positive and negative examples in the second case. These losses were shown to be beneficial over the triplet loss on certain tasks. 
With one positive and multiple negative examples the contrastive loss is defined as:
\begin{equation}
L_{CL} = -\sum_{a \in I}\log \frac{\exp{(s^a \cdot z^a/\tau)}}{ D(a) },
\end{equation}
similarly to our triplet loss formulation, we use as an anchor an embedding of a 3D sketch. As a positive example we use an embedding of a matching 3D shape.
% we use as an anchor an embedding of a 3D sketch and as a positive example an embedding of a matching 3D shape.
$\tau$ is a temperature parameter set to $0.1$ in our experiments, empirically found to give the best results on the validation set. 
% $\tau$ is a temperature parameter set to $0.1$, empirically found to give the best results on the validation set. 

We experimented with multiple choices of the denominator $D(a)$, and found the one which includes distance between the encodings of sketches and shapes, sketches, and shapes gives the best results:
\begin{multline}
    D(a) = \sum_{b \in I }\exp{(s^a \cdot z^b/\tau)} + {\sum_{b \in I \setminus a} \exp{(s^a \cdot s^b/\tau)}  } +\\
    + {\sum_{b \in I \setminus a} \exp{(z^a \cdot z^b/\tau)}  }.
\end{multline}

\section{Experiments and Analysis} 
%\todo{this needs reshuffling to reflect intro -- discussions needed!}
Our backbone architecture consists of an encoder with shared weights for sketches and shapes, trained with the triplet loss.
We then study the choice of an encoder, an architecture and a loss, which allows us to achieve the best performance on our proving ground application -- fine-grained retrieval. We also investigate the effect of human learning, differences in styles and level of details on the retrieval performance.    
%Finally, we offer more generic conclusions by comparing the synthetic sketches with human ones, and demonstrating the advantage of 3D sketches over 2D ones. 
Finally, we compare the synthetic sketches with human ones, and demonstrate the advantage of 3D sketches over 2D ones. 

We evaluate all the methods in terms of Top-$k$ accuracy ($A@k$),
defined as a percentage of queries, which have their ground-truth shapes within k closest retrieval results.
Our test set consists of 202 aligned human sketches (Sec.~\ref{sec:data_split}), and we perform retrieval from 5,794 chair shapes in ShapeNet dataset \cite{shapenet}, excluding the shapes used in the training and validation sets. All the methods are run three times, and are trained for 300 epochs. We select the run and the epoch, which give the best $A@1$ on the validation set.

We use the following abbreviations in the tables: HS = human sketches, SS = synthetic sketches, CN = curve networks, TL = triplet loss, CL = contrastive loss.
For all the baselines we use a mini-batch size of 6, unless specified differently. 
We use a bold font to highlight the best result in each column, and underline the second best. 
$\#$ indicates an experiment number.
 
\subsection{Network design} 
\label{sec:network_design}

%\subsubsection{Encoder choice}
\paragraph{Encoder choice}
%\label{sec:endcoder_choice}
While PointNet\cite{qi2017pointnet} and PointNet++\cite{qi2017pointnet++} are a frequent choice of encoders, we also compare their performance with the more recent DGCNN \cite{wang2019dynamic}. 
On the sketches with inconsistent orientations DGCNN performs worse than PointNet++ (Table \ref{tab:encoder} \#1 vs.~\#3). 
When the sketches are preliminarily aligned to have a consistent orientation (see Sec.~\ref{subsec:data}), the gap in performance of the considered encoders is smaller (Table \ref{tab:encoder}  \#4 vs.~\#6). 
These experiments show that PointNet++ can be a better choice for inaccurate human sketches and can handle a small number of non-consistently oriented sketches in the training set (Table \ref{tab:encoder}  \#1 vs.~\#4).

\begin{table}[h]
\vspace{-8pt}
%\normalsize
\small
  \resizebox{\linewidth}{!}{%
 \small
\begin{tabular}{rrcl|rrr}
\multicolumn{1}{l|}{}   & \multicolumn{3}{c|}{\textbf{Training}}                          & \multicolumn{3}{c}{\textbf{Test set (202 $\rightarrow$ 5,794  )}}                                               \\ \hline
\multicolumn{1}{l|}{\textbf{\#}} & \multicolumn{1}{l|}{\textbf{Size}} & \multicolumn{1}{l|}{\textbf{Data}}                                                    & \textbf{Method}                                                & \multicolumn{1}{c}{\textbf{A@1}} & \multicolumn{1}{c}{\textbf{A@5}} & \multicolumn{1}{c}{\textbf{A@10}} \\ \hline
\multicolumn{1}{r|}{1}           & \multicolumn{1}{r|}{702}           & \multicolumn{1}{c|}{HS ShapeNet}                                                      & \begin{tabular}[c]{@{}l@{}}PointNet++\\ Siam. TL\end{tabular}  & \textbf{26.2}                      & \textbf{43.1}                      & \textbf{54.5}                       \\ \hline
\multicolumn{1}{r|}{2}           & \multicolumn{1}{r|}{702}           & \multicolumn{1}{c|}{HS ShapeNet}                                                      & \begin{tabular}[c]{@{}l@{}}PointNet++\\ Heter. TL\end{tabular} & 19.8                               & 35.2                               & 47.5                                \\ \hline
\multicolumn{1}{r|}{3a}           & \multicolumn{1}{r|}{702}           & \multicolumn{1}{c|}{HS ShapeNet}                                                      & \begin{tabular}[c]{@{}l@{}}DGCNN\\ Siam. TL\end{tabular}                                                     & 20.3                               & 33.2                               & 40.1                                \\ \hline
\multicolumn{1}{r|}{3b}           & \multicolumn{1}{r|}{702}           & \multicolumn{1}{c|}{HS ShapeNet}                                                      & \begin{tabular}[c]{@{}l@{}}DGCNN\\ Siam. TL\end{tabular}                                                     & 22.3                               & 33.2                               & 37.1                                \\ \hline
\multicolumn{1}{r|}{4}           & \multicolumn{1}{r|}{702}           & \multicolumn{1}{c|}{\begin{tabular}[c]{@{}c@{}}HS ShapeNet \\ (aligned)\end{tabular}} & \begin{tabular}[c]{@{}l@{}}PointNet++\\ Siam. TL\end{tabular}  & {\ul 24.8}                         & {\ul 41.6}                         & 48.0                                \\ \hline
\multicolumn{1}{r|}{5}           & \multicolumn{1}{r|}{702}           & \multicolumn{1}{c|}{\begin{tabular}[c]{@{}c@{}}HS ShapeNet \\ (aligned)\end{tabular}} & \begin{tabular}[c]{@{}l@{}}PointNet++\\ Heter. TL\end{tabular} & 18.3                               & 37.6                               & {\ul 48.5}                          \\ \hline
\multicolumn{1}{r|}{6}           & \multicolumn{1}{r|}{702}           & \multicolumn{1}{c|}{\begin{tabular}[c]{@{}c@{}}HS ShapeNet \\ (aligned)\end{tabular}} &\begin{tabular}[c]{@{}l@{}}DGCNN\\ Siam. TL\end{tabular}  & 22.8                               & 35.6                               & 46.5                                \\ \hline
\end{tabular}
}
\vspace{1pt}
\caption{
The evaluation of encoders and architectures: siamese vs.~heterogeneous. 
%See Sec.~\ref{sec:endcoder_choice} and \ref{sec:sim_heter} for the details. 
See Sec.~\ref{sec:network_design} for the details. 
For all the baselines in this table we use a mini-batch size of 6, apart from $\#3b$, where the mini-batch size is equal to 24.
% We use the following abbreviation: HS = human sketches, SS = synthetic sketches, TL = triplet loss.
% For all the baselines in this table we use a mini-batch size of 6. We use an underscore to highlight 2 best results within each column, and a bold font to highlight the best one in each column. $\#$ indicates an experiment number.
}
\label{tab:encoder}
\vspace{-8pt}
\end{table}

%\subsubsection{Siamese vs. Heterogeneous}
\vspace{-8pt}
\paragraph{Siamese vs. Heterogeneous}
%\label{sec:sim_heter}

We compare the Siamese architecture (the sketch and shape encoders share weights) with a heterogeneous architecture (the sketch and shape encoders do not share weights).
% \old{Intuitively, the heterogeneous architecture should behave better due to the obvious differences between shape point clouds and sketches.}
% \new{It is difficult to obtain via a heterogeneous training a strong alignment of feature vectors from two different representations of the same instance, as the alignment is driven only by a triplet loss.}
We observed that if training from scratch, the heterogeneous architecture was not able to converge. We thus initialize the training with the weights of the encoder from the Siamese architecture at the $100$th epoch.
Experiments \#1 vs.~\#2, and \#4 vs.~\#5 in Table \ref{tab:encoder} indicate that heterogeneous training is inferior to Siamese on our training data. 
For Siamese training to work well, one needs to ensure that two different representations of the same instance can be described by the same embedding network. 
In [52], it was observed that when the PointNet encoder is trained on 3D shapes only, there is a minimal subset of points from the original 3D shape point set (a critical point set), such that the embedding of this subset is the same as the embedding of the original 3D shape point set.   We observe that the critical point sets resemble abstraction of 3D shapes with sparse lines in human VR sketches. 
Therefore, Simaese training results in a single efficient encoder for 3D shapes and sketches, which additionally accounts for how humans represent a 3D shape with sparse lines.

%, plus a sparse set of points representing the shape surface. 

%Sketches largely are ridge lines \cite{cole2008people}, plus additional lines that are needed to indicate shape curvature \cite{gryaditskaya2019opensketch}. 
%In the rest of the experiments we use a Siamese architecture where the weights of the sketch and shape encoders are shared.

% \begin{figure}[t]
% % \vspace{-5pt}
% \begin{center}
% \includegraphics[width = \linewidth]{images/dataset_size.pdf}
% \end{center}
% \vspace{-5pt}
%   \caption{ The retrieval accuracy vs. percentage of used human sketches. To obtain these curves we used PointNet++ encoder with TL. The checkpoint was chosen based on best A@1 (a.) or best A@5 (b.) on the validation set.}
% \label{fig:dataset_size}
% \vspace{-10pt}
% \end{figure}

\begin{figure*}[t]
\vspace{-10pt}
\begin{center}
\includegraphics[width = \linewidth]{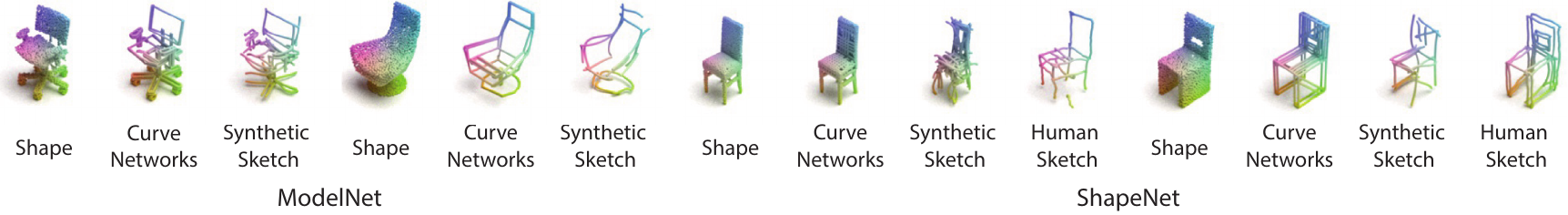}
\end{center}
\vspace{-8pt}
   \caption{Visual comparison of human sketches, with synthetic sketches and curve networks on two shapes datasets: ModelNet and ShapeNet. The curve network extraction step is very sensitive to the quality of the input mesh, naturally it expects a mesh to be a manifold curvature aligned quad mesh, what requires heavy time-consuming mesh processing, where most of readily available remeshing tools either break the shape manifoldness or are unable to process complicated input.}
\label{fig:HS_vs_SS}
\vspace{-10pt}
\end{figure*}

\begin{table}[t]
\vspace{-0pt}
\normalsize
\small
  \resizebox{\linewidth}{!}{%
\begin{tabular}{rrcl|rrr}
\multicolumn{1}{l|}{}   & \multicolumn{3}{c|}{\textbf{Training}}                           & \multicolumn{3}{c}{\textbf{Test set (202 $\rightarrow$ 5,794  )}}                                               \\ \hline
\multicolumn{1}{l|}{\textbf{\#}} & \multicolumn{1}{l|}{\textbf{Size}} & \multicolumn{1}{l|}{\textbf{Data}} & \textbf{Method}                                               & \multicolumn{1}{l}{\textbf{A@1}} & \multicolumn{1}{l}{\textbf{A@5}} & \multicolumn{1}{l}{\textbf{A@10}} \\ \hline
\multicolumn{1}{r|}{1}           & \multicolumn{1}{r|}{702}           & \multicolumn{1}{c|}{HS ShapeNet}   & \begin{tabular}[c]{@{}l@{}}PointNet++\\ Siam. TL\end{tabular} & \textbf{26.2}                      & \textbf{43.1}                      & \textbf{54.5}                       \\ \hline
\multicolumn{1}{r|}{7}           & \multicolumn{1}{r|}{702}           & \multicolumn{1}{c|}{HS ShapeNet}   & \begin{tabular}[c]{@{}l@{}}PointNet++\\ Siam. CL\end{tabular} & 8.9                                & 23.8                               & 31.2                                \\ \hline
\multicolumn{1}{r|}{8}           & \multicolumn{1}{r|}{702}           & \multicolumn{1}{c|}{HS ShapeNet}   & \begin{tabular}[c]{@{}l@{}}DGCNN \\ Siam. CL\end{tabular}     & {\ul 22.3}                         & {\ul 42.1}                         & {\ul 49.5}                          \\ \hline
\end{tabular}
}
\vspace{0.5pt}
\caption{The evaluation of compared losses. See Sec.~\ref{sec:network_design} 
%\ref{sec:losses_comapre} 
for the details. 
For all the experiments we use a mini-batch size of 6, apart from the experiment \#8, where the mini-batch size is set to 24.
}
\label{tab:CL_vs_TL}
\vspace{-10pt}
\end{table}

\vspace{-10pt}
\paragraph{Losses comparison}
%\label{sec:losses_comapre}
Contrastive loss in \cite{khosla2020supervised} is shown to be dependent on the mini-batch size. Due to a GPU memory limitation our mini-batch size is limited to 6 shape-sketch pairs in case of PointNet++. 
%\edit{Due to a GPU memory limitation our batch size is limited to 6 sketch-shape pairs in case of PointNet++.} 
DGCNN has less parameters and we are able to run the training with 24 sketch-shape pairs per mini-batch. 
%\edit{DGCNN has less parameters and we are able to run the training with 9 pairs per batch.} 
In this case it can be seen in Table \ref{tab:CL_vs_TL}  \#7 vs.~\#8 that training with DGCNN is advantageous over training with PointNet++ when using the contrastive loss. Nevertheless, all the results with the contrastive loss lose in performance to the triplet loss, due to a limited batch size. 

% , which is not able to reach its potential due to a limited batch size. 

\subsection{Training data}

\subsubsection{Training data size}
\label{sec:training_data_size}

We analyze how the retrieval performance varies depending on the number of available human sketches.  
% For this experiment, we generate $x\%$ sketch-shape pairs of the full training set. We select sketches randomly, but we keep the ratio between amount of sketches by each of participants roughly constant. 
For this experiment, we randomly select $x\%$ sketch-shape pairs of the full training set.
For each $x\%$ we run an experiment
\setlength{\columnsep}{8pt}%
\begin{wrapfigure}[14]{l}{4.7 cm}
\centering
\vspace{-10pt}
\hspace{-8pt}
\includegraphics[width=4.1cm]{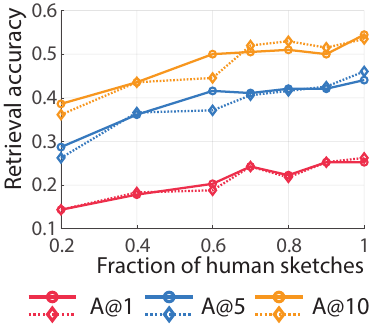}
% \vspace{}
   \caption{ Retrieval accuracy vs.~percentage of human sketches. The checkpoint was chosen based on best A@1 (solid lines) or best A@5 (dashed lines) on the validation set.}
\label{fig:dataset_size}
% \vspace{-10pt}
\end{wrapfigure}
three times with different random selection of sketches. In these experiments we used PointNet++ encoder and TL. We select the checkpoint which gives the best results on the validation set, which is fixed for all the experiments in this paper. It can be seen in Fig.~\ref{fig:dataset_size} that the most rapid improvement due to adding more sketches happens up to $60\%$ of all human sketches in our training set. When we use between $80\%$ to all the sketches the retrieval accuracy starts to stabilize, indicating that simply adding more data might be not enough to boost the performance.

\subsubsection{Human sketches vs Synthetic sketches}
\label{sec:HS_vs_SS}
We compare the training on \emph{human sketches (HS)}, with training on \emph{curve networks (CN)} extracted from 3D shapes with \cite{gori2017flowrep} and on \emph{synthetic sketches (SS)}, obtained from curve networks using the code from \cite{Luo2020VRSketch}. We generate curve networks and synthetic sketches for the set of shapes for which we collected human sketches. Since the method \cite{gori2017flowrep} expects the mesh to be an edge-manifold mesh, we process all the meshes using the algorithm by Huang et al.~\cite{huang2020manifoldplus}, followed by a decimation step to reduce meshes size.  We additionally use randomly picked 702 SSs for a subset of ModelNet shapes, provided by \cite{Luo2020VRSketch}. Fig.~\ref{fig:HS_vs_SS} shows a visual comparison between HSs, CNs and SSs.
%humans sketches, curve networks and synthetic sketches. 
It can be seen in Table \ref{tab:HS_vs_SS} \#1 vs.~\#9-11 that training with human sketches marginally outperforms training with either of SSs or CNs. 

\begin{table}[h]
\vspace{-5pt}
\center
\footnotesize
%\resizebox{\linewidth}{!}{%
\begin{tabular}{rrc|rrr}
\multicolumn{1}{l|}{}   & \multicolumn{2}{c|}{\textbf{Training}}                          & \multicolumn{3}{c}{\textbf{Test set (202 $\rightarrow$ 5,794  )}}                    \\ \hline
\multicolumn{1}{l|}{\#} & \multicolumn{1}{l|}{\textbf{Size}} & \multicolumn{1}{l|}{\textbf{Data}}                     & \multicolumn{1}{c}{\textbf{A@1}} & \multicolumn{1}{c}{\textbf{A@5}} & \multicolumn{1}{c}{\textbf{A@10}} \\ \hline
\multicolumn{1}{r|}{1}  & \multicolumn{1}{r|}{702}  & HS ShapeNet                                   & \textbf{26.2}           & \textbf{43.1}           & \textbf{54.5}            \\ \hline
\multicolumn{1}{r|}{9}  & \multicolumn{1}{r|}{702}  & SS ModelNet                                   & 11.9                    & { \ul 25.7}             & {\ul 32.7}               \\ \hline
\multicolumn{1}{r|}{10} & \multicolumn{1}{r|}{702}  & CN ShapeNet                                   & 7.4                     & 21.3                    & 26.2                     \\ \hline
\multicolumn{1}{r|}{11} & \multicolumn{1}{r|}{702}  & SS ShapeNet                                   & {\ul 13.4}              & 25.3                    & 30.7                     \\ \hline
\end{tabular}%
%}
\vspace{2pt}
\caption{Human sketches vs.~synthetic sketches.  All the networks in this table use PointNet++ \cite{qi2017pointnet++} as an encoder, and are trained with the Triplet Loss and Siamese architecture.
%See Sec.~\ref{sec:HS_vs_SS} for the details.
}
\label{tab:HS_vs_SS}
\vspace{-6pt}
\end{table}

\vspace{-4pt}
\subsubsection{Augmentation on human sketches}
\label{sec:augment}
The previous section concludes that the retrieval accuracy on human sketches when training only on synthetic sketches falls short of the retrieval accuracy when the training is done on human sketches, However, we observe that synthetic sketches can be used in addition to human sketches to boost the performance of the fine-grained retrieval: Table \ref{tab:augment} (\#12 vs.~\#l, and \#13 vs.~\#4) shows that $A@5/10$ are consistently improved.
% Despite that using only synthetic sketches the performance falls short of the accuracy when the training is done on human sketches, we observe that synthetic sketches can be used in addition to human sketches to boost the performance of the fine-grained retrieval: Table \ref{tab:augment} (\#12 vs.~\#l, and \#13 vs.~\#4) shows that $A@5/10$ are consistently improved.

\begin{table}[h]
% \vspace{-6pt}
\footnotesize
%  \resizebox{\linewidth}{!}{%
\begin{tabular}{rrc|rrr}
\multicolumn{1}{l|}{}   & \multicolumn{2}{c|}{\textbf{Training data size and type}}                          & \multicolumn{3}{c}{\textbf{Test set (202 $\rightarrow$ 5,794  )}}                                               \\ \hline
\multicolumn{1}{l|}{\textbf{\#}} & \multicolumn{1}{l|}{\begin{tabular}[c]{@{}r@{}} \textbf{Sketches/} \\ \textbf{shapes}\end{tabular}}                                       & \multicolumn{1}{c|}{\textbf{Data}}                                                                                                          & \multicolumn{1}{c}{\textbf{A@1}} & \multicolumn{1}{c}{\textbf{A@5}} & \multicolumn{1}{c}{\textbf{A@10}} \\ \hline
\multicolumn{1}{r|}{1}           & \multicolumn{1}{r|}{702 / 702}                                                 & \multicolumn{1}{c|}{HS ShapeNet}                                                              &   {  26.2}                      & {43.1}                        & 54.5                                \\ \hline
\multicolumn{1}{r|}{4}           & \multicolumn{1}{r|}{702 /702 }                                           & \multicolumn{1}{c|}{\begin{tabular}[c]{@{}c@{}}HS ShapeNet  (aligned)\end{tabular}}                 & {24.8}                         & 41.6                               & 48.0                                \\ \hline
\multicolumn{1}{r|}{12}          & \multicolumn{1}{r|}{\begin{tabular}[c]{@{}r@{}}702+702 /\\ 702+702\end{tabular}}  & \multicolumn{1}{c|}{\begin{tabular}[c]{@{}c@{}} HS ShapeNet \\ + SS ModelNet\end{tabular}}      & 24.3                               & \textbf{48.0}                      & \textbf{55.9}                       \\ \hline
\multicolumn{1}{r|}{13}          & \multicolumn{1}{r|}{\begin{tabular}[c]{@{}r@{}}702+702 /\\ 702+702\end{tabular}} & \multicolumn{1}{c|}{\begin{tabular}[c]{@{}c@{}} HS ShapeNet (aligned) \\ +SS ModelNet\end{tabular}}  & 24.3                               & {  44.1}                         & 53.0               \\ \hline  
\multicolumn{1}{r|}{14}          & \multicolumn{1}{r|}{\begin{tabular}[c]{@{}r@{}}702+Aug./\\ 702\end{tabular}} & \multicolumn{1}{c|}{\begin{tabular}[c]{@{}c@{}}HS ShapeNet\\ +Aug. distortions\end{tabular}}  & {\textbf{28.2}}                         & {  44.1}                         & { 55.0}                          \\ \hline
\multicolumn{1}{r|}{15}          & \multicolumn{1}{r|}{\begin{tabular}[c]{@{}r@{}}702+Aug./ \\ 702\end{tabular}} & \multicolumn{1}{c|}{\begin{tabular}[c]{@{}c@{}}HS ShapeNet\\ +Aug. rotations\end{tabular}}      & 19.8                               & 35.6                               & 45.1                      \\ \hline          
\multicolumn{1}{r|}{\AgHSRep}          & \multicolumn{1}{r|}{\begin{tabular}[c]{@{}r@{}}702+363 /\\ 702\end{tabular}} & \multicolumn{1}{c|}{\begin{tabular}[c]{@{}c@{}}HS ShapeNet\\ + Aug. style\end{tabular}}      & 24.3                               & 45.5                               & 54.0                      \\ \hline          
\multicolumn{1}{r|}{\AgSSRep}          & \multicolumn{1}{r|}{\begin{tabular}[c]{@{}r@{}}702+363 / \\702 \end{tabular}} & \multicolumn{1}{c|}{\begin{tabular}[c]{@{}c@{}}HS ShapeNet\\ + SS ShapeNet\end{tabular}}      & 24.3                               & 41.6                               & 49.0                      \\ \hline          
\end{tabular}
%}
\vspace{1pt}
\caption{The evaluation of augmentation strategies. All the networks in this table use PointNet++ \cite{qi2017pointnet++} as an encoder, and are trained with the Triplet Loss and Siamese architecture. 
}
\label{tab:augment}
\vspace{-8pt}
\end{table}

Table \ref{tab:augment} (\#14, vs.~\#12 and vs.~\#1) shows that a simple data augmentation of human sketches by applying random distortions along x,y,z-axes can lead to a higher top-1 accuracy than training on the combination of human sketches and synthetic sketches.
Despite that PointNet++ performs better on a non-aligned dataset (Table \ref{tab:encoder} \#1 vs.~\#4), when we apply rotations on sketches as an augmentation strategy the performance drops (Table \ref{tab:augment} \#15). This implies that in the practical retrieval application either users should provide the orientation of a sketch or the retrieval system should address this problem explicitly.

Additionally, we evaluate how the fine-grained retrieval performance changes if we provide for each shape more than one sketch. In the experiment \AgHSRep{}, we train on 702 sketches from our training set augmented with 363 sketches by one of the participants from group B, providing a second sketch representation for 363 3D shapes.  In the experiment \AgSSRep{}, we instead augment with 363 synthetic sketches of the same shapes. It can be seen that having several shapes drawn by difference participants does not bring much (Table \ref{tab:augment} \#16 vs.~\#1). Using synthetic sketches gives a negative effect on the retrieval performance  (Table \ref{tab:augment} \#17 vs.~\#1). This allows to drive out an important conclusion: \emph{When collecting a dataset one should aim at maximizing the diversity of shapes and styles rather than aiming at having several shapes drawn by several different participants.}

To further understand the effect of data augmentation with synthetic sketches, we study augmentation when only $40\%$ of our training set of human sketches is used (Table \ref{tab:aug_detailed}), and vary the number of synthetic sketches. We observe that augmenting with synthetic sketches improves the $A@1/5/10$ till the number of synthetic sketches is 1.5 to 2.0 times higher than the number of human sketches. We thus conclude that a certain number of synthetic sketches can help increase the retrieval accuracy, but the ratio between human and synthetic sketches should be taken into account. Adding more synthetic sketches results in performance degradation. Augmentation by sketches distortions still achieves the highest $A@1$ (last line in Table \ref{tab:aug_detailed}).

\begin{table}[th]
\footnotesize
%  \resizebox{\linewidth}{!}{%
\begin{tabular}{l|rrr}
                                  & \multicolumn{3}{c}{\textbf{\begin{tabular}[c]{@{}c@{}}Test set \\ (202 $\rightarrow$ 5,794 )\end{tabular}}} \\ \hline
\textbf{Data}                     & \multicolumn{1}{c}{\textbf{A@1}}         & \multicolumn{1}{c}{\textbf{A@5}}         & \multicolumn{1}{c}{\textbf{A@10}}         \\ \hline
281 HS ShapeNet                   & 17.82                             & 36.14                             & 43.56                              \\ \hline
281 HS ShapeNet + 140 SS ModelNet & 19.31                             & 36.14                             & 43.56                              \\
281 HS ShapeNet + 280 SS ModelNet & {\ul 20.79}                       & 36.14                             & 45.54                              \\
281 HS ShapeNet + 421 SS ModelNet & 20.30                             & \textbf{43.07}                    & {\ul 49.50}               \\
281 HS ShapeNet + 561 SS ModelNet & 20.30                             & {\ul 39.60}                       & \textbf{50.00}                     \\
281 HS ShapeNet + 702 SS ModelNet & 19.80                             & 36.63                             & 48.02                              \\ \hline
281 HS ShapeNet + Aug. distortions                         & \textbf{21.78}                    & 38.61                             & 44.06                              \\ \hline
\end{tabular}
\vspace{1pt}
\caption{The detailed evaluation of augmentation strategies.}
\label{tab:aug_detailed}
\vspace{-10pt}
%}
\end{table}

%As in the previous experiments, augmentation with distortions allows achieving the highest $A@1$.

\begin{table*}[t]
\vspace{-4pt}
\normalsize
\small
  \resizebox{\linewidth}{!}{%
\begin{tabular}{lrcl|rrr|rrr|rrr|rrr}
\multicolumn{1}{l|}{}   & \multicolumn{3}{c|}{\textbf{Training}}                         & \multicolumn{3}{c|}{\textbf{\begin{tabular}[c]{@{}c@{}}Test set \\ (202 $\rightarrow$ 5,794  )\end{tabular}}}    & \multicolumn{3}{c}{\textbf{\begin{tabular}[c]{@{}c@{}}Unseen 5 participants \\ (50 $\rightarrow$ 5,794 ) (U)\end{tabular}}} & \multicolumn{3}{c|}{\textbf{\begin{tabular}[c]{@{}c@{}}Group A \textbackslash U \\ (67 $\rightarrow$ 5,794  )\end{tabular}}} & \multicolumn{3}{c}{\textbf{\begin{tabular}[c]{@{}c@{}}Group B \\ (85 $\rightarrow$ 5,794  )\end{tabular}}}      \\ \hline
\multicolumn{1}{l|}{\textbf{\#}} & \multicolumn{1}{l|}{\textbf{Size}}                                       & \multicolumn{1}{l|}{\textbf{Data}}                                                            & \textbf{Method}                                               & \multicolumn{1}{l}{\textbf{A@1}} & \multicolumn{1}{l}{\textbf{A@5}} & \multicolumn{1}{l|}{\textbf{A@10}} & \multicolumn{1}{l}{\textbf{A@1}}     & \multicolumn{1}{l}{\textbf{A@5}}     & \multicolumn{1}{l|}{\textbf{A@10}}    & \multicolumn{1}{l}{\textbf{A@1}}     & \multicolumn{1}{l}{\textbf{A@5}}     & \multicolumn{1}{l|}{\textbf{A@10}}     & \multicolumn{1}{l}{\textbf{A@1}} & \multicolumn{1}{l}{\textbf{A@5}} & \multicolumn{1}{l}{\textbf{A@10}} \\ \hline
\multicolumn{1}{r|}{1}           & \multicolumn{1}{r|}{702}                                                 & \multicolumn{1}{c|}{HS ShapeNet}                                                              & \begin{tabular}[c]{@{}l@{}}PointNet++ Siam. TL\end{tabular} & { 26.2}                         & 43.1                               & 54.5                                 & { 24.0}                             & { 38.0}                             & 44.0                                    & { 11.9}                             & 22.4                                   & { 35.8}                               & { 38.8}                         & { 62.4}                         & {75.3}                       \\ \hline
\end{tabular}
}
\vspace{-1pt}
\caption{Results analysis per group. 
%See Sec.~\ref{sec:per_group} for the details. 
}
\label{tab:per_group}
\vspace{-4pt}
\end{table*}

\subsection{Generalization across users}
\label{sec:per_group}

As described in Sec.~\ref{sec:data_split} we have two groups of participants: group A -- those who each sketched 10 shapes and group B -- those who sketched more. 
It can be seen in Fig.~\ref{fig:sketches_split} that sketches in group B are more accurate and have more details, and the retrieval performance on such sketches is higher than average, as shown in Table \ref{tab:per_group}. 
The sketches of unseen participants contain moderate amount of details (Fig.~\ref{fig:sketches_split}), and the performance on those sketches is slightly below the average performance. This indicates that the methods generalize sufficiently well to unseen styles. 
Following these results, the retrieval accuracy for the sketches of overlapping 129 shapes by one of the participants (ID = 38) from group B is higher than for the sketches of participants from group A:  $31.78\%/58.14\%/68.99\%$ versus $13.95\%/33.33\%/44.96\%$ $A@1/5/10$. These 129 sketches are not a part of a test set of 202 sketches.

\subsection{2D sketch vs.~3D sketch}\label{sec:2D_vs_3D}

\begin{table}[h]
\centering
\resizebox{\columnwidth}{!}{%
\small
\begin{tabular}{l|l|l|lll}
\# & Method                                                                                       & Data                                                                   & A@1            & A@5            & A@10           \\ \hline
A  & \begin{tabular}[c]{@{}l@{}}2d$\rightarrow$3d:   \\ Multi-view,  Heter. TL\end{tabular}        & 2D HS + 3D shapes                                                      & 19.8           & 45.05          & 59.41          \\
B  & 2d$\rightarrow$3d:   PointNet++ TL                                                           & 2D HS + 3D shapes                                                      & 11.39          & 34.65          & 54.95          \\ \hline
C  & \begin{tabular}[c]{@{}l@{}}(3 views) 2d$\rightarrow$3d: \\ Multi-view,  Heter. TL \end{tabular}           & \begin{tabular}[c]{@{}l@{}}2D HS (3 views) \\ + 3D shapes\end{tabular} & 32.67          & 68.81          & 80.69          \\ \hline
D  & 2d$\rightarrow$2d:   \cite{yu2016sketch}                                                                    & 2D HS + 3D views                                                       & 25.25          & 60.4           & 77.23          \\
E  & 2d$\rightarrow$2d:   \cite{yu2016sketch}                                                                    & 2D HS + 2D NPR                                                         & 14.36          & 45.54          & 60.89          \\ \hline
1  & \begin{tabular}[c]{@{}l@{}} 3d$\rightarrow$3d: \\ PointNet++, Siam. TL\end{tabular} & \begin{tabular}[c]{@{}l@{}}3D HS \\ + 3D shapes\end{tabular}           & \textbf{61.39} & \textbf{83.67} & \textbf{90.59} \\ \hline
\end{tabular}%
 }
 \vspace{1pt}
\caption{The comparison of 2D sketch-based and 3D sketch-based retrieval methods. See Sec.~\ref{sec:2D_vs_3D} for the details. }
\label{tab:2D_vs_3D}
\vspace{-8pt}
\end{table}

%As discussed in the introduction, \cite{qi_2021} is the only work we are aware of that attempts to solve the problem of fine-grained retrieval from a single 2D sketch. For each 3D shape, they collect 2D sketches by novices from 3 predefined viewpoints: They show one reference image to a volunteer for 15 seconds, then display a blank canvas and let the volunteer sketch just observed object from memory using fingers on a tablet/phone. In our work we use the same set of shapes, but a different split to training and test data. In their case the training set consists of 804 sketches and test set of 201 sketches. We refer the interested reader to their work for an evaluation of diverse baselines, and only compare with their results using the numbers provided in their work. Since in their case the gallery consists of 201 shapes, matching the shapes in the test set, we evaluate all results in this section on a gallery consisting of 202 shapes from the test set (Sec.~\ref{sec:data_split}). Additionally we compare with a baseline, where we use all three available 3D sketch views as input to the network. We use a Siamese architecture based on NGVNN \cite{he2019view}, following the experiments in \cite{qi_2021}. 
%As it can be seen in Table \ref{tab:2D_vs_3D} \#16, \#17 3D sketch-based retrieval outperforms by far the 2D sketch-based one, indicating that a 3D VR sketch has the potential to become a common input for many new practical applications. 

We are not aware of any work on 2D sketch-based \textit{instance-level} 3D shape retrieval.
We thus adopt \textit{category-level} (sketch/image)-based shape retrieval methods, such as \cite{lee2018cross, he2018triplet}, for a fine-grained scenario.
% Unlike many retrieval papers, we use a very large gallery size (5794 shapes) to evaluate the fine-grained aspect. 
Due to a large domain gap between a 2D sketch and a 3D shape, we limit in this section the gallery size from 5794 to 202 shapes from the test set.
We use the dataset of 2D human sketches done by participants without any art experience\footnote{http://sketchx.ai/downloads/: AmateurSketch-3DChair} on the same set of shapes as our dataset. 
We compare two architectures, where the 2D branch is VGG11 \cite{simonyan2014very}\footnote{VGG11 gives better results than ResNet \cite{he2016deep} in our experiments}, and the 3D shape branch either employs
(1) NGVNN \cite{he2019view} with multi-view shape representation (\#A), or (2) PointNet++ with point cloud shape representation (\#B). The training is performed with a Triplet Loss. 
In $\#C$ we combine information from 3 2D sketch views of the same shape with azimuth angles of 0,30 and 75$^o$. 
Finally, we compare with a 2D sketch to a 2D image fine-grained retrieval baseline \cite{yu2016sketch}. 
In $\#D$ we use as a target a shape view (Phong shading) from the same viewpoint as a sketch. 
In $\#E$ we use as a target a 2D NPR rendering from the same viewpoint as a sketch. $\#C, D, E$ use encoders pretrained on ImageNet.
The numerical evaluation in Table \ref{tab:2D_vs_3D} demonstrates that \emph{3D sketches can bring a new level of accuracy to \textit{instance-level} retrieval.}

%
%
%We compare 3D sketch-based retrieval with 2D sketch based. 
%
%
%
%As it can be seen in Table \ref{tab:2D_vs_3D} \#16, \#17 3D sketch-based retrieval outperforms by far the 2D sketch-based one, indicating that a 3D VR sketch has the potential to become a common input for many new practical applications. 
%For visual comparison please refer to the supplemental.
%\todo{more discussions needed here for sure, especially we make a big deal out of this in intro}

% \begin{figure*}
% % \vspace{-5pt}
% \begin{center}
% \includegraphics[width = \linewidth]{images/retrieval_results_v2.pdf}
% \end{center}
% \vspace{-5pt}
%   \caption{The retrieval results for the training on human sketches with the triplet loss and Siamese architecture (Experiment #1). For visual result of other methods please refer to the supplemental. We roughly divide the results in this figure in three groups. \emph{Good}: The retrieval results where the ground-truth shape is within top-5 retrieval results. \emph{Medium}: The retrieval results where the retrieved shapes look visually similar to the ground-truth shape, but due to the distortions present in human sketches, the ground-truth is not among top ranked retrieval results. \emph{Failure cases}: The retrieval results, where the top ranked results differ strongly from the ground-truth shapes.} \todo
% \label{fig:results}
% \vspace{-10pt}
% \end{figure*}
\begin{figure*}
% \vspace{-5pt}
\begin{center}
\includegraphics[width = \linewidth]{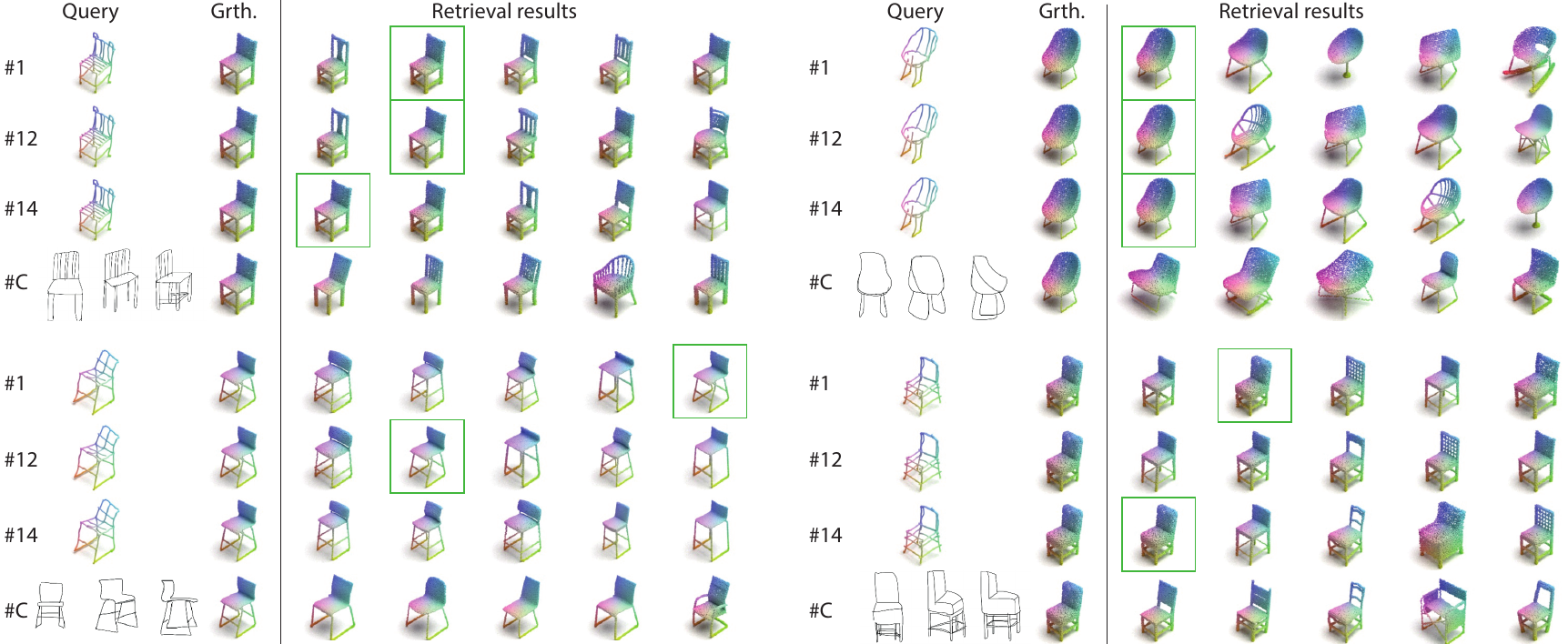}
\end{center}
\vspace{-7pt}
   \caption{The ranked retrieval results from 5,794 shapes when training on 702 human sketches with the triplet loss and Siamese architecture (\#1), with augmentation by synthetic sketches of additional 702 shapes (\#12), with augmentation with random distortions (\#14). The 3D retrieval results are compared with the retrieval results given 3 2D viewpoints (\#C).
   For more results please refer to the supplemental. The green box shows the position of the ground-truth.} 
\label{fig:results}
\vspace{-10pt}
\end{figure*}

\subsection{Discussion}
\label{sec:discussion}
% \paragraph{Sketching in 3D}
% % When collecting our dataset we were interested to see how people without art education and with diverse set of sketching skills and familiarity with VR will tackle 3D VR sketching. Our questionnaire results indicate that overall participants found sketching in VR satisfying. The main concerns were that the headset is too heavy and some people felt dizzy after 20 minutes of sketching. 
% % The participants, on average, were not fully satisfied with their sketches. 
% We hope that our dataset can help studying sketching in 3D, and formulating novel sketching rules tailored to 3D environment. For instance, it is an interesting question how to denote the curvature of the surface in 3D, some participants opt to draw a grid, others trace parallel lines (Fig.~\ref{fig:sketches_split}). In 2D concept sketching designers use cross sections to denote surface curvature \cite{gryaditskaya2019opensketch}, can such technique be adopted to 3D? Developing simple guidelines on how to maintain sense of dimension and perspective in 3D remains an open question. We hope that our dataset of sketches, complemented with questionnaires and time-space information for each stroke, will contribute to defining sketching strategies, which will consequently lead to more accurate and descriptive shape depictions.   

% \vspace{-10pt}
% \paragraph{Retrieval}
Fig.~\ref{fig:results} shows the visualization of the retrieval results. We observe that the retrieved results frequently match well the input sketches, but do not necessary match the ground-truth -- due to the sketch being a distorted version of the shape. This analysis highlights the need for new structure-aware metrics, more tolerant towards distortions. Similarly, the retrieval method should be structure-aware to account for potential distortions. Nevertheless, all the 3D retrieval solutions manage to grasp the shape from a few sparse strokes, achieving notably higher accuracy than the retrieval results from the comparable 2D sketches.
% \refactor{Fig.~\ref{fig:results} shows the visualization of the retrieval results, we observe the next causes of a poor numerical evaluation: (i) the retrieved results frequently  match well the input sketches, but do not match the ground-truth -- due to a sketch being a distorted version of the shape (e,h); (ii) similarly, when sketches are drawn not accurately enough, the retrieved results will resemble the less-accurate sketch (f,g); (iii) certain non-common shapes are not learned well (i,k), but we also observe many successful retrieval result for rare shapes; (iv) network can not disambiguate shape features from lines indicating solid surfaces, and in many cases captures well the overall shape, but not the details (j,l).}

% \todo{rewrite this part changing to the insights on the dataset collection}
% \refactor{Since collecting human sketches is a labor-intensive and time-consuming task, we believe designing appropriate data augmentation strategies could constitute sensible future work. Our analysis of the results in the previous section demonstrate the clear gap between synthetic data \cite{Luo2020VRSketch} and real world data, calling for novel NPR algorithms. One such direction can be to learn sketch-like appearance from the available collected data, or to learn a distribution of displacement vectors with respect to a reference 3D shape.}

% Finally, our experiments with the contrastive loss indicate the need for light-weight encoders to provide more options for novel losses designs. 

\section{Conclusion} 
We propose the first large-scale dataset of human VR sketches to fulfill a vision on the synergy between sketching and VR.
With the dataset, we (i) demonstrate the advantage of 3D sketches over 2D sketches for navigating large 3D shapes collections, and (ii) analyze the role of training set size and alternative augmentation strategies.

Our experiments suggest that to better address the remaining domain gap between 3D VR sketch and 3D shapes, future work should focus on how to better account for structural information, adapting to the distortions inherent to human sketches. Such algorithms will further require deformation-aware losses. 
Our experiments with the contrastive loss indicate the need for light-weight encoders to provide more options for novel losses designs. 
Nevertheless, carefully leveraging existing tools, we demonstrate notable improvement in instance-level retrieval accuracy over the previous work \cite{Luo2020VRSketch}. Our best method in terms of Top-1 accuracy: PointNet++ encoder trained with a triplet loss on human sketches with augmentation by distortions, achieves $A@1 = 28.2$, $A@5 = 44.1$, and $A@10 = 55.0$, while the best model from \cite{Luo2020VRSketch}, trained as was proposed in their work, on our data reaches just  $A@1 = 2.5$, $A@5 = 5.5$,  $A@10 = 8.4$. 
Through our experiments we conclude that (i) when collecting a fine-grained dataset of VR sketches one should aim at maximizing the diversity of shapes and styles rather than aiming at having several shapes drawn by several different participants, (ii) the synthetic sketches can be used as data augmentation when they provide additional shapes diversity and their number does not exceed the number of human sketches, and (iii) a training set of 600-700 sketches provides a good estimate of the achievable performance of the fine-grained retrieval method.
Our dataset is the first step towards adoption of 3D VR tools by an average consumer. We believe it will foster research on multiple topics, such as design of novel encoder and decoder architectures, and loss designs for retrieval and reconstruction problems. 
% \old{We plan to release our dataset, and the custom-build VR sketching interface to facilitate future research.}
Our dataset and the custom-build VR sketching interface are available at \url{tinyurl.com/VRSketch3DV21}.

\clearpage
{\small
\bibliographystyle{ieee_fullname}
\bibliography{camera_ready}
}

\end{document}